\documentclass{article}
\usepackage{spconf,amsmath,graphicx}
\usepackage{amssymb,amsfonts}
\usepackage[normalem]{ulem}
\usepackage{graphicx}
\usepackage{float}
\useunder{\uline}{\ul}{}

\title{Style Adaptation for Domain-adaptive Semantic Segmentation}
\name{Ting Li\textsuperscript{1,2,3}, Jianshu Chao\textsuperscript{1,2,}*, Deyu An\textsuperscript{1,2}}
\address{\textsuperscript{1}Quanzhou Institute of Equipment Manufacturing, Fujian Institute of Research on the Structure\\ 
of Matter, Chinese Academy of Sciences\\
\textsuperscript{2}Fujian College, University of Chinese Academy of Sciences\\
\textsuperscript{3}College of Computer and Cyber Security, Fujian Normal University}
\begin{document}
%
\maketitle
\begin{abstract}
Unsupervised Domain Adaptation (UDA) refers to the method that utilizes annotated source domain data and unlabeled target domain data to train a model capable of generalizing to the target domain data. Domain discrepancy leads to a significant decrease in the performance of general network models trained on the source domain data when applied to the target domain. We introduce a straightforward approach to mitigate the domain discrepancy, which necessitates no additional parameter calculations and seamlessly integrates with self-training-based UDA methods. Through the transfer of the target domain style to the source domain in the latent feature space, the model is trained to prioritize the target domain style during the decision-making process. We tackle the problem at both the image-level and shallow feature map level by transferring the style information from the target domain to the source domain data. As a result, we obtain a model that exhibits superior performance on the target domain. Our method yields remarkable enhancements in the state-of-the-art performance for synthetic-to-real UDA tasks. For example, our proposed method attains a noteworthy UDA performance of 76.93 mIoU on the GTA→Cityscapes dataset, representing a notable improvement of +1.03 percentage points over the previous state-of-the-art results.
\end{abstract}
\section{Introduction}
\label{sec:intro}

Neural Networks \cite{gu2018recent} and Transformers \cite{vaswani2017attention} have achieved great success in semantic segmentation tasks, but supervised tasks typically require a large amount of annotated data. Pixel-level annotation is needed, with at least an hour for each image \cite{Cordts2016Cityscapes}, which significantly increases the cost. One approach to address this problem is to utilize existing annotated data or easily obtainable synthetic data to train models and test them on target data. However, due to domain differences, the model's performance metrics often decline substantially when tested on target data. In order to obtain a more robust model, researchers have proposed UDA methods \cite{hoyer2022daformer}\cite{hoyer2022hrda}\cite{hoyer2023mic}, transferring knowledge from annotated source domain data to unannotated target data. 


It has been proven that CNNs are sensitive to distribution shifts \cite{NEURIPS2018_018b59ce} in image classification. Recent studies \cite{9711504} have shown that Transformers are more robust compared to these factors. 
In addition, CNNs mainly focus on texture \cite{geirhos2018imagenettrained}, while Transformers emphasize shape, which is more similar to human vision. Some researches have revealed significant differences between the induction bias of standard CNNs and human vision: humans primarily rely on object content (i.e., shape) for recognition \cite{landau1988importance}, while CNNs exhibit a strong preference for style (i.e., texture) \cite{geirhos2018imagenettrained}. This explains why CNNs are more susceptible to changes when switching between domains, as image style is more likely to vary across different domains. 

Early studies \cite{zhou2021domain}\cite{huang2017arbitrary}\cite{pan2018two} have confirmed that feature distribution shifts caused by style differences mainly occur in the shallow layers of the network. This implies that the shallow layers' feature distribution in the network can reflect the style information of the input images.  Therefore, following these works' methods, we manipulate the style features of the feature maps in the shallow layers of the network. The feature extractor captures the style features of the target domain while preserving the content of the source domain. This approach weakens the style features of the source domain while enhancing the style features of the target domain, achieving style feature transfer. 


\section{Method}
\label{sec:method}
\subsection{Image to Image Domain Adaptation}
\label{sec:img2img}
In UDA, we are given a source dataset as \begin{equation}\begin{aligned}D^s=\{(x_i^s,y_i^s)\}_{i=1}^{N_s}\end{aligned}\end{equation} where $N_s$ is the number of the color images in the dataset, and $\begin{aligned}y^s\in\mathbb{R}^{H\times W}\end{aligned}$ represents the associated semantic map of $\begin{aligned}x^s\in\mathbb{R}^{H\times W\times3}\end{aligned}$. Similarly, \begin{equation}D^t=\{x_i^t\}_{i=1}^{N_t}\end{equation} is the target dataset where true semantic labels are missing. Typically, segmentation networks trained on $D^s$ exhibit performance degradation when tested on $D^t$. Here, we use Fourier Domain Adaptation (FDA) \cite{Yang_2020_CVPR} and RGB adaptation to reduce the domain gap between the two datasets at the image-level.

FDA aims to minimize domain differences by replacing the low-frequency components in the target domain with those from the source domain. This is because low-frequency components can be inferred as the domain style. FDA has achieved significant improvements in semantic segmentation. Therefore, we employ the FDA method for data augmentation, as expressed by the formula:
\begin{equation}x^{s\to t}=\mathcal{F}^{-1}([\beta\circ \mathcal{F}^A(x^t)+(1-\beta)\circ \mathcal{F}^A(x^s),\mathcal{F}^P(x^s)])\label{eq1}\end{equation}

The variables $\mathcal{F}^A$ and $\mathcal{F}^P$ denote the amplitude and phase components of the Fourier transform, respectively. In the inverse Fourier transform, the phase and amplitude components are remapped to the image space. The hyperparameter $\beta$ determines the filter's size in the inverse Fourier transform.

Random RGB shift is a prevalent and widely adopted technique for data augmentation. Through our experimental observations, we fortuitously discovered that employing random RGB shift as a data augmentation technique significantly enhances the model's performance. Our hypothesis is that the image-level implementation of random RGB shift enables a closer resemblance between the style of the source and target domains, thereby mitigating the domain gap. Building upon the concept of random RGB shift, we introduce a RGB adaptation method as a solution for domain adaptation.

The mean value of each channel is calculated for RGB images $x$ as follows:
\begin{equation}\mu(x)=\frac1{HW}\sum_{h=1}^H\sum_{w=1}^Wx_{hw}\label{eq}\end{equation}

\begin{equation}x^{s\to t}=x^s+(\mu(x^t)-\mu(x^s))\label{eq}\end{equation}

The variables $\mu(s)$ and $\mu(t)$ represent the mean values of the source domain image and the target domain image, respectively, along the channel dimension. By employing this method, the content of the source domain image remains unaltered, thus preserving the availability of accurate labels. Additionally, it facilitates the closer alignment of the source domain image with the target domain image within the RGB space. 




\subsection{Style Adaptive Instance Normalization}
In UDA methods, the primary factor causing domain shift is the disparity in styles across domains. The presence of domain shift constrains the models' capacity for generalization in both domain adaptation and domain generalization tasks. Previous studies have demonstrated that shallow features extracted by backbone networks possess the capability to capture style information in images. Established approaches typically characterize the style features of an image by computing the mean and standard deviation along the channel dimension of shallow features. 
\begin{equation}\sigma(x)=\sqrt{\frac1{HW}\sum_{h=1}^{H}\sum_{w=1}^{W}(x_{hw}-\mu(x))^2+\epsilon}\label{eq}\end{equation}

Conventional instance normalization can eliminate specific stylistic information from an image. Directly applying this method to UDA can diminish the network's capacity to learn the style information of the source domain images. However, it also disregards the style information of the target domain, resulting in diminished performance and limited generalization ability on the target domain. 

To decrease the network's ability to learn style information from the source domain images while enhancing the style information of the target domain images, we apply AdaIN \cite{huang2017arbitrary} to replace the style information of the source domain images with that of the target domain images. Meanwhile, this method retains the content information of the source domain images. We term the proposed approach as Style Adaptive Instance Normalization (SAIN). The specific implementation formula is as follows:
\begin{equation}SAIN(x^{s},x^{t})=\sigma(x^{t})\left(\frac{x^{s}-\mu(x^{s})}{\sigma(x^{s})}\right)+\mu(x^{t})\label{eq}\end{equation}

$\mu$ and $\sigma$ represent the mean and standard deviation of the feature map in the channel dimension, respectively. 

By transferring the style of the target domain to the source domain during the training process, the network $g_\theta$ biased towards content no longer relies on the style of the source domain to make decisions but focuses more on content while also paying attention to the style of the target domain. During testing, we directly use network $g_\theta$ without SAIN to ensure the independence of predictions and reduce computational burden. Therefore, we replace the original loss function with a content-biased loss, shown as follows:
\begin{equation}\mathcal{L}_{i}^{S}=-\sum_{j=1}^{H\times W}\sum_{c=1}^{C}y_{(i,j)}^{S}\log SAIN\left(g_{\theta}(x_{i}^{S})^{(j,c)},g_{\theta}(x_{i}^{T})^{(j,c)}\right)\label{eq}\end{equation}

Furthermore, we follow the consistency training in DA-Former, which involves training the teacher network on augmented target data using DACS \cite{tranheden2021dacs}, while the teacher model generates pseudo-labels using non-augmented target images. 


\section{Experiments}
\label{sec:guidelines}

\subsection{Implementation Details}
The proposed method is applied to two challenging unsupervised domain adaptation tasks, where there are abundant semantic segmentation labels in the synthetic domain (source domain), but not in the real domain (target domain). The two synthetic datasets used are GTA5 \cite{richter2016playing} and SYNTHIA \cite{ros2016synthia}, while the real domain dataset is CityScapes \cite{Cordts2016Cityscapes}. 




The proposed method is validated based on the DAFormer network and the Mix Transformer-B5 encoder \cite{xie2021segformer}. All backbone networks are pretrained on ImageNet. In the default UDA setting, the MIC \cite{hoyer2023mic} masked image self-training strategy and the training parameters are used, including the AdamW optimizer, the encoder learning rate of $6\times10^{-5}$, the decoder learning rate of $6\times10^{-4}$, 60k training iterations, a batch size of 2, linear learning rate warm-up, 
and DACS \cite{tranheden2021dacs} data augmentation.

\subsection{Evaluation}
First, we integrate RGB adaptation with several significant UDA methods, including DAFormer \cite{hoyer2022daformer}, HRDA \cite{hoyer2022hrda} and MIC \cite{hoyer2023mic}, using the DAFormer framework. Table 1 demonstrates that RGB adaptation achieves notable improvement compared to previous UDA methods without RGB adaptation. 

Karras et al. \cite{karras2019style} demonstrated that styles at different levels encode distinct visual attributes. Styles from fine-grained spatial resolution (lower levels in our network) encode low-level attributes like color and fine textures, whereas styles from coarse-grained spatial resolution (higher levels in our network) encode high-level attributes including global structure and textures. Therefore, the application of our SAIN module at the appropriate level is necessary to mitigate adverse style-induced biases. The networks from Block 1 to Block 4 become increasingly deeper. Figure 1 illustrates that the most notable improvement is achieved when applying SAIN in Block 3. However, applying SAIN to features at excessively low levels only has a limited impact on reducing feature biases. Additionally, using SAIN in excessively high-level styles may result in the loss of essential semantic information. Through our experimental findings, we discovered that the concurrent application of SAIN to both Block 2 and Block 3 results in optimal performance.

Visual comparisons are conducted with the second performer (i.e., MIC), which utilizes the same segmentation network backbone as ours. Figure 2 illustrates that our model's prediction results demonstrate higher accuracy. Additionally, our approach demonstrates strong performance on some common categories, including the first row with the terrain, wall in the second row and building in the third and truck in fourth rows. We attribute this phenomenon to the transferability of RGB adaptation and SAIN, which enables the model to learn more style information from the target domain. 

\begin{table}[]
\caption{Performance (IoU) of RGB adaptation with different UDA methods on GTA→Cityscapes.}
\label{table}
\setlength{\tabcolsep}{2pt}
\begin{tabular}{c|cccc}
\hline
Network  & UDA Method & w/o RGB Adapt. & w/ RGB Adapt.  \\ \hline
DAFormer & DAFormer   & 68.3           & 69.37          \\
DAFormer & HRDA       & 73.8           & 74.45          \\
DAFormer & MIC        & 75.9           & 76.64          \\ \hline
\end{tabular}
\end{table}

\begin{figure}[!t]
\centerline{\includegraphics[width=\columnwidth]{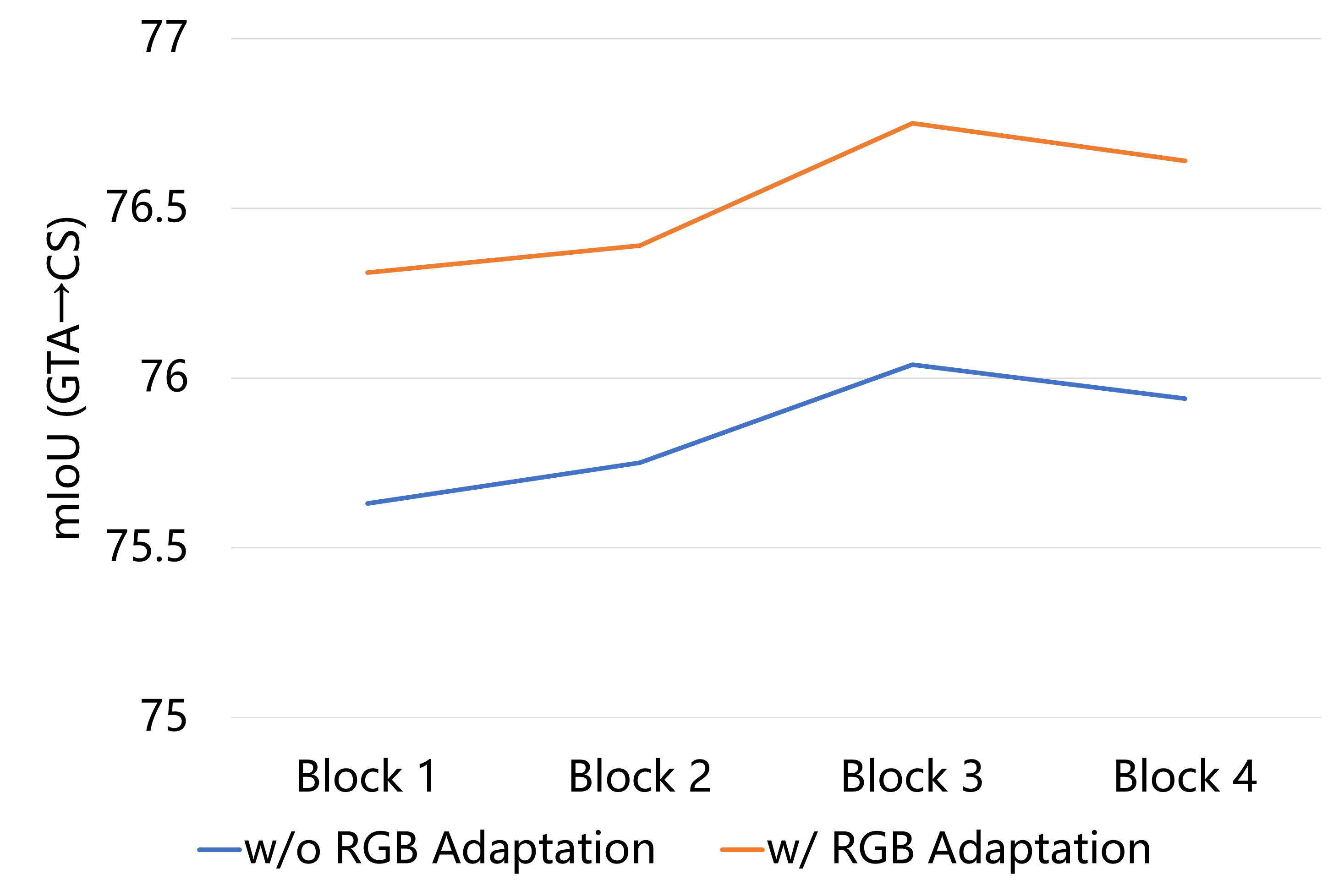}}
\caption{The effect of SAIN on different blocks.}
\label{fig2}
\end{figure}

\subsection{Influence of Style on UDA}
In the following, we analyze the underlying principles of our method on GTA→Cityscapes. Firstly, we analyze the impact of SAIN on UDA at various feature levels. As shown in Figure 1, as the network depth increases from Block 1 to Block 3, the improvement in the performance of UDA using SAIN also increases accordingly. The results in Table 2 and Table 3 demonstrate significant performance improvements across all benchmarks. In particular, our method has led to a +1.03 increase in mIoU for GTA→CS and a +1.05 increase for Synthia→CS. 
For most categories, such as building, fence, rider, truck, and train, there is a certain performance improvement. However, there are also some categories that have a slight performance decrease after using SAIN, such as bike. This may be due to the difference in annotation strategies for the bike category between the Cityscapes dataset and the GTA dataset. 

\begin{table*}[]
\caption{Semantic segmentation performance (IoU) on GTA→Cityscapes}
\label{table}
\setlength{\tabcolsep}{2pt}
\resizebox{\textwidth}{!}{
\begin{tabular}{c|ccccccccccccccccccc|c}
\hline
Method &
  Road &
  S.walk &
  Build. &
  Wall &
  Fence &
  Pole &
  Tr.light &
  Tr.sign &
  Veget. &
  Terrain &
  Sky &
  Person &
  Rider &
  Car &
  Truck &
  Bus &
  Train &
  M.bike &
  Bike &
  mIoU \\ \hline
ADVENT &
  89.4 &
  33.1 &
  81.0 &
  26.6 &
  26.8 &
  27.2 &
  33.5 &
  24.7 &
  83.9 &
  36.7 &
  78.8 &
  58.7 &
  30.5 &
  84.8 &
  38.5 &
  44.5 &
  1.7 &
  31.6 &
  32.4 &
  45.5 \\
DACS &
  89.9 &
  39.7 &
  87.9 &
  30.7 &
  39.5 &
  38.5 &
  46.4 &
  52.8 &
  88.0 &
  44.0 &
  88.8 &
  67.2 &
  35.8 &
  84.5 &
  45.7 &
  50.2 &
  0.0 &
  27.3 &
  34.0 &
  52.1 \\
ProDA &
  87.8 &
  56.0 &
  79.7 &
  46.3 &
  44.8 &
  45.6 &
  53.5 &
  53.5 &
  88.6 &
  45.2 &
  82.1 &
  70.7 &
  39.2 &
  88.8 &
  45.5 &
  59.4 &
  1.0 &
  48.9 &
  56.4 &
  57.5 \\
DAFormer &
  95.7 &
  70.2 &
  89.4 &
  53.5 &
  48.1 &
  49.6 &
  55.8 &
  59.4 &
  89.9 &
  47.9 &
  92.5 &
  72.2 &
  44.7 &
  92.3 &
  74.5 &
  78.2 &
  65.1 &
  55.9 &
  61.8 &
  68.3 \\
HRDA &
  96.4 &
  74.4 &
  91.0 &
  \textbf{61.6} &
  51.5 &
  57.1 &
  63.9 &
  69.3 &
  91.3 &
  48.4 &
  {\ul 94.2} &
  79.0 &
  52.9 &
  93.9 &
  84.1 &
  85.7 &
  75.9 &
  63.9 &
  {\ul 67.5} &
  73.8 \\
MIC &
  \textbf{97.4} &
  \textbf{80.1} &
  {\ul 91.7} &
  61.2 &
  {\ul 56.9} &
  {\ul 59.7} &
  {\ul 66.0} &
  {\ul 71.3} &
  {\ul 91.7} &
  {\ul 51.4} &
  \textbf{94.3} &
  {\ul 79.8} &
  {\ul 56.1} &
  {\ul 94.6} &
  {\ul 85.4} &
  {\ul 90.3} &
  {\ul 80.4} &
  {\ul 64.5} &
  \textbf{68.5} &
  {\ul 75.9} \\ \hline
Ours &
  {\ul 97.24} &
  {\ul 79.12} &
  \textbf{92.15} &
  {\ul 61.45} &
  \textbf{58.5} &
  \textbf{60.98} &
  \textbf{69.23} &
  \textbf{72.58} &
  \textbf{91.93} &
  \textbf{53.33} &
  93.99 &
  \textbf{81.26} &
  \textbf{60.68} &
  \textbf{94.84} &
  \textbf{88.3} &
  \textbf{90.5} &
  \textbf{83.24} &
  \textbf{65.59} &
  66.82 &
  \textbf{76.93} \\ \hline
\end{tabular}
}
\label{tab1}
\end{table*}

\begin{table*}[]
\caption{Semantic segmentation performance (IoU) on Synthia→Cityscapes. }
\label{table}
\setlength{\tabcolsep}{2pt}
\resizebox{\textwidth}{!}{
\begin{tabular}{c|ccccccccccccccccccc|c}
\hline
Method &
  Road &
  S.walk &
  Build. &
  Wall &
  Fence &
  Pole &
  Tr.Light &
  Tr.Sign &
  Veget. &
  Terrain &
  Sky &
  Person &
  Rider &
  Car &
  Truck &
  Bus &
  Train &
  M.bike &
  Bike &
  mIoU \\ \hline
ADVENT &
  85.6 &
  42.2 &
  79.7 &
  8.7 &
  0.4 &
  25.9 &
  5.4 &
  8.1 &
  80.4 &
  – &
  84.1 &
  57.9 &
  23.8 &
  73.3 &
  – &
  36.4 &
  – &
  14.2 &
  33.0 &
  41.2 \\
DACS &
  80.6 &
  25.1 &
  81.9 &
  21.5 &
  2.9 &
  37.2 &
  22.7 &
  24.0 &
  83.7 &
  – &
  90.8 &
  67.6 &
  38.3 &
  82.9 &
  – &
  38.9 &
  – &
  28.5 &
  47.6 &
  48.3 \\
ProDA &
  {\ul 87.8} &
  45.7 &
  84.6 &
  37.1 &
  0.6 &
  44.0 &
  54.6 &
  37.0 &
  \textbf{88.1} &
  – &
  84.4 &
  74.2 &
  24.3 &
  88.2 &
  – &
  51.1 &
  – &
  40.5 &
  45.6 &
  55.5 \\
DAFormer &
  84.5 &
  40.7 &
  88.4 &
  41.5 &
  6.5 &
  50.0 &
  55.0 &
  54.6 &
  86.0 &
  – &
  89.8 &
  73.2 &
  48.2 &
  87.2 &
  – &
  53.2 &
  – &
  53.9 &
  61.7 &
  60.9 \\
HRDA &
  85.2 &
  47.7 &
  88.8 &
  {\ul 49.5} &
  4.8 &
  57.2 &
  65.7 &
  60.9 &
  85.3 &
  – &
  92.9 &
  79.4 &
  52.8 &
  89.0 &
  – &
  \textbf{64.7} &
  – &
  63.9 &
  {\ul 64.9} &
  65.8 \\
MIC &
  86.6 &
  {\ul 50.5} &
  {\ul 89.3} &
  47.9 &
  {\ul 7.8} &
  {\ul 59.4} &
  {\ul 66.7} &
  {\ul 63.4} &
  {\ul 87.1} &
  – &
  {\ul 94.6} &
  {\ul 81.0} &
  {\ul 58.9} &
  {\ul 90.1} &
  – &
  {\ul 61.9} &
  – &
  {\ul 67.1} &
  64.3 &
  {\ul 67.3} \\
Ours &
  \textbf{89.06} &
  \textbf{57.39} &
  \textbf{90.1} &
  \textbf{51.37} &
  \textbf{7.99} &
  \textbf{60.53} &
  \textbf{69.03} &
  \textbf{63.44} &
  86.57 &
  – &
  \textbf{94.91} &
  \textbf{82.33} &
  \textbf{61.1} &
  \textbf{89.4} &
  – &
  57.28 &
  – &
  \textbf{67.92} &
  \textbf{65.24} &
  \textbf{68.35} \\ \hline
\end{tabular}
}
\end{table*}

\begin{figure*}[!t]
\centerline{\includegraphics[width=2\columnwidth]{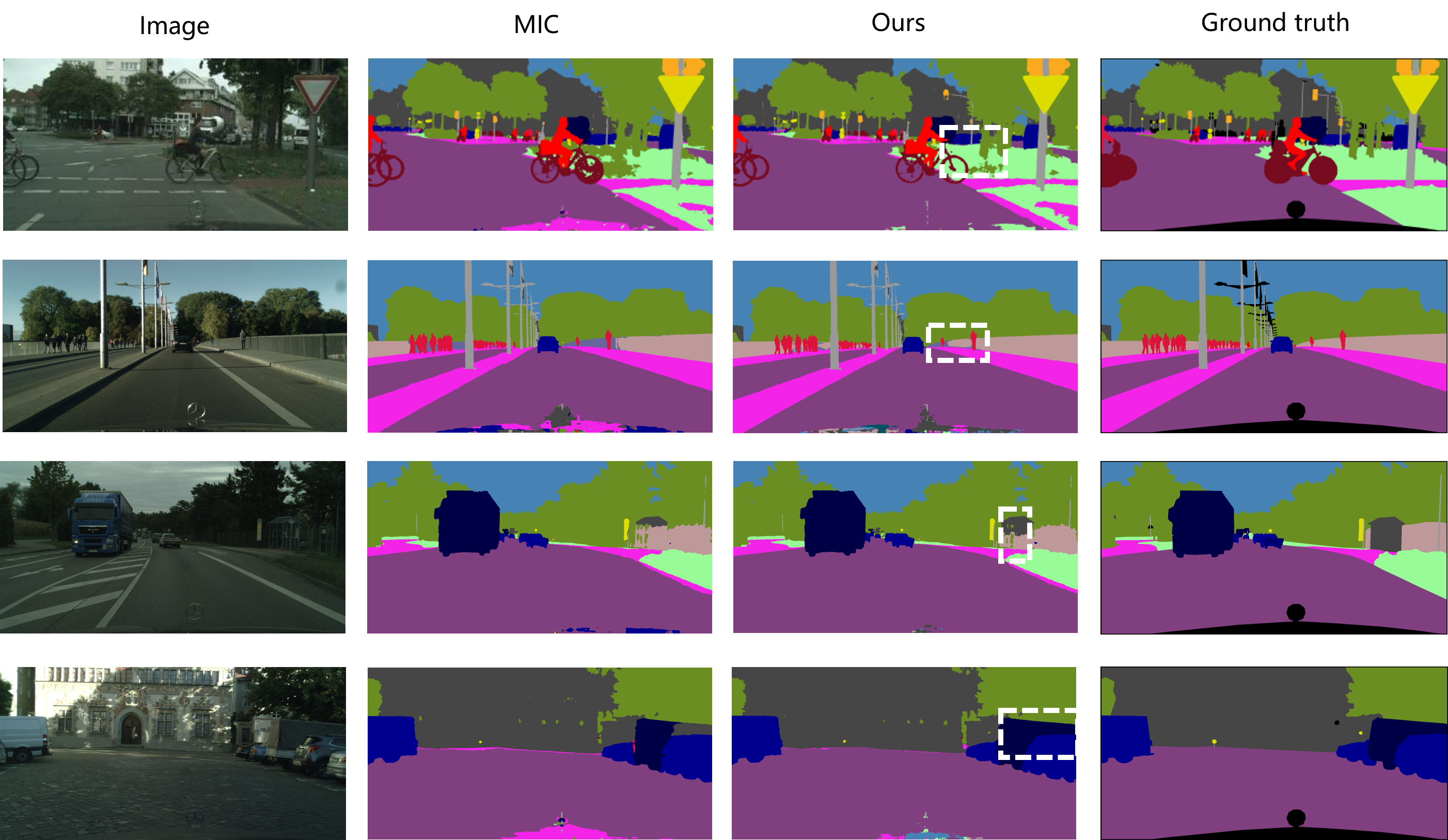}}
\caption{Qualitative comparison with the previous state-of-the-art method MIC on GTA→CS. The proposed method gets better segmentation for classes such as terrain, fence, building, and truck.}
\label{seg1}
\end{figure*}

\section{Conclusion}
We have proposed a straightforward method for reducing domain discrepancy, which requires no additional learning and can be seamlessly integrated into self-supervised UDA. By transferring the target domain style to the source domain within the latent feature space, the model is trained to prioritize the style of the target domain during its decision-making process. Our experiments validate the remarkable performance enhancements achieved by our proposed method in Transformer-based domain adaptation. Despite its simplicity, the results indicate that our method actually surpasses the current state-of-the-art techniques. This suggests that the distributional misalignment caused by shallow-level statistics can indeed impact cross-domain generalization, but it can be mitigated through image translation and SAIN. The issue of model robustness in machine learning remains a challenging problem, and while we do not assert that our method is optimal, its simplicity may also yield performance improvements in other domain adaptation tasks.

\noindent\textbf{Acknowledgements:} This work is supported by STS Project of Fujian Science and Technology Program (No. 2023T3042). 




\clearpage
\vfill\pagebreak
\bibliographystyle{IEEEbib}
\bibliography{refs}

\end{document}